\documentclass[10pt,twocolumn,letterpaper]{article}

\usepackage[pagenumbers]{cvpr} 

\usepackage{graphicx}
\usepackage{amsmath}
\usepackage{amssymb}
\usepackage{booktabs}
\usepackage[accsupp]{axessibility}

\usepackage{xcolor}
\newcommand{\jf}[1]{{\textcolor{black}{#1}}}

\usepackage[pagebackref,breaklinks,colorlinks]{hyperref}
\usepackage[accsupp]{axessibility}  

\usepackage[capitalize]{cleveref}
\crefname{section}{Sec.}{Secs.}
\Crefname{section}{Section}{Sections}
\Crefname{table}{Table}{Tables}
\crefname{table}{Tab.}{Tabs.}

\def\cvprPaperID{1411} 
\def\confName{CVPR}
\def\confYear{2023}

\begin{document}

\title{Being Comes from Not-being: \\Open-vocabulary Text-to-Motion Generation with Wordless Training}



\author{Junfan Lin$^{1,2}$ \ Jianlong Chang$^{3}$ \  Lingbo Liu$^{2}$ \  Guanbin Li$^{1}$ \  Liang Lin$^{1}$ \  Qi Tian$^{3}$\thanks{Corresponding author: tian.qi1@huawei.com} \ \ Chang Wen Chen$^{2}$\\
{\tt\small $^1$Sun Yat-sen University \ \  $^2$The Hong Kong Polytechnic University \ \ $^3$Huawei Cloud}
}

\maketitle

\begin{abstract}
Text-to-motion generation is an emerging and challenging problem,  which aims to synthesize motion with the same semantics as the input text. However, due to the lack of diverse labeled training data, most approaches either limit to specific types of text annotations or require online optimizations to cater to the texts during inference at the cost of efficiency and stability. In this paper, we investigate offline open-vocabulary text-to-motion generation in a zero-shot learning manner that neither requires paired training data nor extra online optimization to adapt for unseen texts. Inspired by the prompt learning in NLP, we pretrain a motion generator that learns to reconstruct the full motion from the masked motion. During inference, instead of changing the motion generator, our method reformulates the input text into a masked motion as the prompt for the motion generator to ``reconstruct'' the motion. In constructing the prompt, the unmasked poses of the prompt are synthesized by a text-to-pose generator. To supervise the optimization of the text-to-pose generator, we propose the first text-pose alignment model for measuring the alignment between texts and 3D poses. And to prevent the pose generator from overfitting to limited training texts, we further propose a novel wordless training mechanism that optimizes the text-to-pose generator without any training texts. The comprehensive experimental results show that our method obtains a significant improvement against the baseline methods. The code is available at \url{https://github.com/junfanlin/oohmg}.
\end{abstract}


\section{Introduction}
\label{sec:intro}

\begin{figure}[t]
	\centering
	\begin{center}
		\centering
		\includegraphics[clip=True, trim={10 15 25 0}, width=\linewidth]{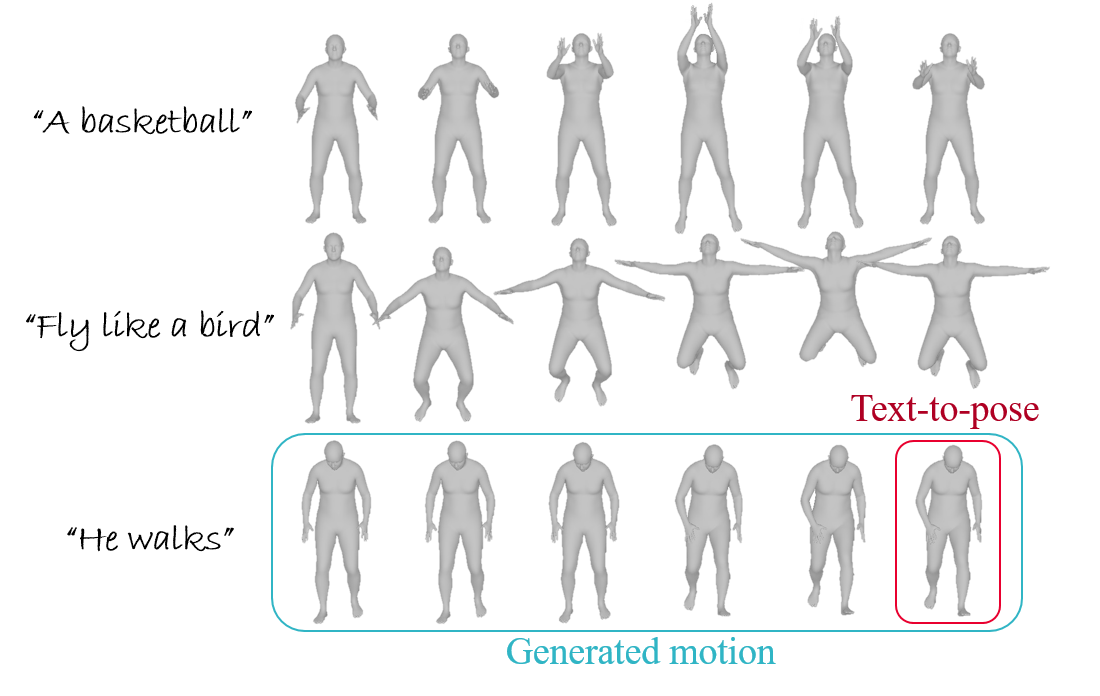}
	\end{center}
	\vspace{-5mm}
	\caption{Demonstrations of our OOHMG. Given an unseen open-vocabulary text (e.g., an object name ``a basketball", or a simile description ``fly like a bird", or a usual text ``he walks"), OOHMG translates the text into the text-consistent pose, which is used to prompt the motion generator for synthesizing the motion.}
	\vspace{-3mm}
	\label{fig:first_page}
\end{figure}
Motion generation has attracted increasing attention due to its practical value in the fields of virtual reality, video games, and movies. Especially for text-conditional motion generation, it can largely improve the user experience if the virtual avatars can react to the communication texts in real time. However, most current text-to-motion approaches are trained on paired text-motion data with limited types of annotations, and thus could not well-generalize to unseen open-vocabulary texts. 

To handle the open-vocabulary texts, recent works leverage the powerful zero-shot text-image alignment ability of the pretrained model, i.e., CLIP~\cite{radford2021learning}, to facilitate the text-to-motion generation. Some works like MotionCLIP~\cite{tevet2022motionclip} use the CLIP text encoder to extract text features and learn a motion decoder to decode the features into motions. However, they require paired text-motion training data and still could not handle texts that are dissimilar to the training texts. Instead of learning an offline motion generator with paired data, some works like AvatarCLIP~\cite{hong2022avatarclip} generate motions for the given textual descriptions via online matching and optimization. Nevertheless, matching cannot generate new poses to fit diverse texts and online optimization is usually time-consuming and unstable.

\begin{figure}[t]
	\centering
	\begin{center}
		\includegraphics[clip=True, trim={25, 20, 0, 0}, width=\linewidth]{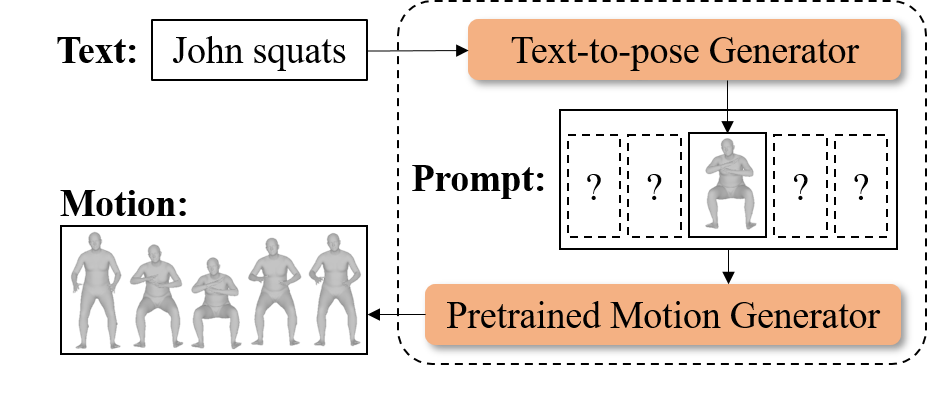}
	\end{center}
	\vspace{-6mm}
	\caption{The sketch of OOHMG. A text is fed to the text-to-pose generator to obtain a text-consistent pose. Then, the pose is used to construct the motion prompt for the pretrained motion model to generate a motion.}
	\vspace{-3mm}
	\label{fig:overview}
\end{figure}
In this paper, we investigate filling the blank of offline open-vocabulary text-to-motion generation in a zero-shot learning manner. For convenience, we term our method as \textbf{OOHMG} which stands for \textbf{O}ffline \textbf{O}pen-vocabulary \textbf{H}uman \textbf{M}otion \textbf{G}eneration. The main philosophy of OOHMG is inspired by prompt learning~\cite{sun2022paradigm,brown2020language,liu2022prompt,yu2022towards,wang2023lion,wang2022fine} in the field of natural language processing (NLP). Specifically, instead of changing the pretrained motion generator to cater to the given texts online, OOHMG reformulates the texts into a familiar input format to prompt the pretrained motion generator for synthesizing motions in the manner of ``reconstruction". As for prompt construction, OOHMG learns a text-to-pose generator using the novel wordless training mechanism so that the pose generator can generalize to unseen texts during inference. After training, OOHMG uses the text-to-pose generator to translate texts into poses to construct the prompt. The overall sketch and demonstrations of OOHMG are illustrated in Fig.~\ref{fig:overview} and Fig.~\ref{fig:first_page}, respectively. In this sense, the two key ingredients of OOHMG include the motion generator pretraining and the prompt construction for open-vocabulary texts. In the following, we further elaborate on each of these ingredients.


As for the motion generator, we learn a motion generator by mask-reconstruction self-supervised learning. Particularly, our method adopts a bidirectional transformer-based~\cite{vaswani2017attention} architecture for the motion generator. During training, the motion generator takes the randomly-masked motions as inputs and is optimized to reconstruct the original motions. To predict and reconstruct the masked poses from the unmasked, the motion generator is required to focus on learning motion dynamics which is the general need for diverse motion generation tasks. By this means, unlike previous methods that design different models for different tasks~\cite{GunjanAggarwal2021Dance2MusicAD,lin2018human,HyeminAhn2018Text2ActionGA,ChuanGuo2020Action2MotionCG}, our motion model can be directly applied to diverse downstream tasks by unifying the input of these tasks into masked motions to prompt the generator for motion generation. Moreover, our generator can flexibly control the generated content, such as the number, the order, and the positions of different poses of the generated motion by editing the masked motions, resulting in a controllable and flexible motion generation.




In constructing the motion prompt for open-vocabulary motion generation, OOHMG learns a text-to-pose generator and uses it to generate the unmasked poses of the masked motions, as shown in Fig.~\ref{fig:overview}. There are two major difficulties in learning the text-to-pose generator: 1) what can associate diverse texts and poses to supervise the pose generator, and 2) how to obtain diverse texts as the training inputs. For difficulty 1, we build the first large-scale text-pose alignment model based on CLIP, namely TPA, that can efficiently measure the alignment between texts and 3D SMPL poses~\cite{LeonidPishchulin2017BuildingSS,SMPL-X:2019} in the feature space. 
With TPA, the text-to-pose generator learns to generate poses for texts by maximizing the text-pose alignments via gradient descent. As for difficulty 2, instead of collecting massive texts laboriously for training, we consider an extreme training paradigm, termed wordless training. Just as its name implies, wordless training only samples random training inputs from the latent space of texts. And we found that the optimized pose generator can well-generalize to real-world texts.

Overall, the contributions of OOHMG are as follows. 1) We propose an offline open-vocabulary text-to-motion generation framework, inspired by prompt learning, and 2) to supervise the training process of the text-to-pose generator, we propose the first text-pose alignment model, i.e., TPA, and 3) to endow the text-to-pose generator with the ability to handle open-vocabulary texts, we train the generator with the novel wordless training mechanism. 4) Extensive experiment results show that OOHMG is able to generate motions for open-vocabulary texts efficiently and effectively, and obtain clear improvement over the advanced baseline methods qualitatively and quantitatively.

\section{Related Work}
\textbf{Conditional Motion Generation} \jf{can be classified into various categories based on the types of conditions. For example, music has been utilized as a condition in some studies to generate dance motions~\cite{GunjanAggarwal2021Dance2MusicAD}, while others have synthesized movements through short motion descriptions~\cite{lin2018human,ChaitanyaAhuja2019Language2PoseNL,HyeminAhn2018Text2ActionGA} and action labels~\cite{ChuanGuo2020Action2MotionCG,MathisPetrovich2021ActionConditioned3H}. The success of these methods is heavily dependent on large motion capture datasets~\cite{cai2022humman,cai2021playing,CatalinIonescu2014Human36MLS,NaureenMahmood2019AMASSAO,DushyantMehta2016Monocular3H,GlVarol2017LearningFS,TimovonMarcard2018RecoveringA} and labeled motion description datasets, including AMASS~\cite{AbhinandaRPunnakkal2021BABELBA}, KIT motion-language dataset~\cite{MatthiasPlappert2016TheKM}, and HumanML3D dataset~\cite{Guo_2022_CVPR}. However, such datasets are often limited by their task design and data collection challenges, such as the failure to account for emotional movements. Although several methods have demonstrated impressive qualitative and quantitative results~\cite{TEACH:3DV:2022,zhang2022motiondiffuse}, those trained on limited datasets are unable to generalize to open-vocabulary motion descriptions.}

\jf{\textbf{Probing Knowledge from Pretrained Model.} The development of pretrained foundation models has led to the potential for zero-shot/few-shot learning to surpass supervised learning~\cite{AlecRadford2021LearningTV,devlin2018bert,brown2020language,wang2022mvsnet}. One such model, CLIP~\cite{AlecRadford2021LearningTV}, has the ability to semantically align language-vision latent spaces~\cite{YaelVinker2022CLIPassoSO}. Combined with CLIP, DALL-E~\cite{radford2021learning} enables impressive text-to-image synthesis capabilities. This powerful representation ability of foundation model has led to the emergence of zero-shot text-driven applications~\cite{KevinFrans2021CLIPDrawET,OrPatashnik2021StyleCLIPTM,peng2021neural,huang2022audio}, including 3D meshes generation~\cite{jain2022zero,NikolayJetchev2022ClipMatrixTC,OscarMichel2022Text2MeshTN,AdityaSanghi2021CLIPForgeTZ,peng2021animatable}.}

\jf{Related to ours, recent studies~\cite{RobinRombach2022HighResolutionIS,ChitwanSaharia2022PhotorealisticTD} have combined CLIP with diffusion generation models to generate text-consistent 3D meshes~\cite{poole2022dreamfusion,jain2022zero}, while other methods focus on generating static meshes or 2D images for text-video generation, e.g. Make-A-Video~\cite{singer2022make}, Imagen Video~\cite{ho2022imagen}, and Phenaki~\cite{villegas2022phenaki}. As for open-vocabulary motion generation, CLIP-Actor~\cite{youwang2022clip} simply uses motion from existing datasets by matching the textual descriptions with the motion labels of the existing text-motion datasets. And MotionCLIP~\cite{tevet2022motionclip} learns a motion VAE by regularizing the latent space to align with the feature space of CLIP, which also requires labeled data. AvatarCLIP~\cite{hong2022avatarclip} is the closest method to ours, as it also explores the potential of zero-shot open-vocabulary motion generation, but our approach does not require online matching or optimization.}

\section{Preliminaries}

\jf{This paper investigates offline zero-shot open-vocabulary human motion generation (OOHMG). To address open-vocabulary texts, OOHMG includes a text-pose alignment model based on the text-image alignment model, i.e., CLIP~\cite{radford2021learning}. In this section, we provide a brief introduction to the task as well as CLIP.}

\jf{\textbf{Open-vocabulary 3D human motion generation} involves generating a motion $m$ that aligns with a given natural language motion description $d$, such as "fly like a bird." A motion is a sequence of 3D poses, $m=[p_t]_{t=1:T}$, where $p$ is the 3D pose, $t$ represents the timestep, and $T$ is the maximum length of the motion. We use a 6D-rotation representation $p\in \mathbb{R}^{J\times 6}$~\cite{zhou2019continuity}, but we also utilize the latent representation $p^l\in \mathbb{R}^{32}$ of VPoser~\cite{SMPL-X:2019} which is a well-known pose VAE trained on massive poses, to incorporate the pose prior from VPoser. Our focus is on generating body motion, and therefore we do not consider facial expressions, hand poses, or global orientation. We utilize SMPL~\cite{MatthewLoper2015SMPLAS}, a popular parametric human model, for its interpretability and compatibility with various platforms. SMPL is a parametric human model driven by large-scale aligned human surface scans~\cite{LeonidPishchulin2017BuildingSS}, and feeding pose representations into SMPL enables us to obtain 3D meshes $v=\mathcal{M}_\text{SMPL}(p)$.}

\jf{\textbf{CLIP}~\cite{radford2021learning} is a vision-language pre-trained model designed for large-scale image-text datasets. It comprises two encoders: an image encoder $E_o$ and a text encoder $E_{d}$. We use $o$ to denote the image and $d$ to represent the text. The encoders are trained such that the latent codes of paired images and texts are pulled together, while the unpaired codes are pushed apart. Formally, the CLIP loss function is 
\begin{align}
\mathcal{L}_{\text{CLIP}}(\mathbf{o}, \mathbf{d}) = -\sum_{i=1:B} \log \textbf{Pr}(o_i|d_i) - \log \textbf{Pr}(d_i|o_i), \label{equ:cliploss}
\end{align}
where $\mathbf{o}$ and $\mathbf{d}$ is the sets of images $\{o_i\}_{i=1:B}$ and texts $\{d_i\}_{i=1:B}$, and $B$ is the batch size. $\textbf{Pr}$ is the softmax probability of the $o_i$ given $d_i$ in a batch, \jf{vice versa}. Particularly, to calculate $\textbf{Pr}(o_j|d_i)$, the cosine similarity between text feature $E_o(o_i)$ and each image feature $E_d(d_j)$ of the batch data is calculated, and the temperature-softmax operation is applied to the cosine similarities. Formally, 
\begin{align}
\textbf{Pr}(o_i|d_i) = \frac{\exp\Big(\text{cossim}\big(E_o(o_i), E_d(d_i)\big)/H\Big)}{\underset{j=1:B}{\sum}{\exp\Big(\text{cossim}\big(E_o(o_j), E_d(d_i)\big)/H\Big)}},
\end{align}
where $\text{cossim}(f_i, f_j) = \frac{f_i ^T f_j}{|f_i||f_j|}$, and $H$ is the temperature to adjust the sensitivity of softmax. The calculation of $\textbf{Pr}(d_i|o_i)$ follows the similar process. 
For convenience, we use \textbf{CLIP score} to stand for the cosine similarity between text and image features from CLIP.}

\section{OOHMG}

As mentioned above, our OOHMG achieves offline open-vocabulary text-to-motion generation with two key ingredients, i.e., the pretrained motion generator and prompt construction. Both of these components manage to be text-free during the training phase. In this section, we detail these two modules formally.

\begin{figure}[t]
	\begin{center}
		\includegraphics[clip=True, trim={0, 0, 0, 0}, width=\linewidth]{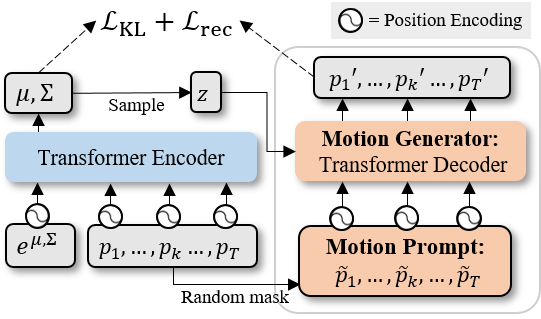}
	\end{center}
	\vspace{-6mm}
	\caption{Training process of motion generator. The motion encoder takes in the original motion and extracts the latent feature. The motion generator takes in masked motion and the latent feature to reconstruct the original motion.}
	\label{fig:pose2motion}
	\vspace{-3mm}
\end{figure}
\subsection{Motion Generator Pretraining}
In advanced language modeling in NLP, the language model~\cite{devlin2018bert} learns to reconstruct the masked sentence from the randomly masked sentence in a self-supervised manner. Our motion generator also follows a similar training strategy. Specifically, during training, the random proportion of the poses of a motion $m = [p_t]_{t=1:T}$ are masked by a learnable embedding $e^\text{mask}\in \mathcal{R}^{|p|}$. Formally, the pose $\tilde{p}_t$ of the masked motion $\tilde{m}$ is generated by $\tilde{p}_t = c_t \times p_t + (1 - c_t)\times e^\text{mask}$, where $c_t\in\{0, 1\}$ is a binary random condition sampled for each timestep $t\in[1,T]$. When $c_t$ equals 1, the original pose is preserved. Otherwise, the pose is replaced by the mask embedding. In addition, since there is usually more than one motion corresponding to the same masked motion, to prevent learning a generator that generates the average motions, we also adopt a motion encoder to extract the latent feature for each motion. The motion encoder follows ACTOR~\cite{MathisPetrovich2021ActionConditioned3H} without the motion category conditions, as illustrated in Fig.~\ref{fig:pose2motion}. The motion generator takes in $\tilde{m}$ and the latent code from the motion encoder to predict the $m'$ to reconstruct the original $m$. \jf{Different from ACTOR which is optimized to encode and decode the full sequence, ours is to predict the full sequence from the masked sequence. Hence, to meet different requirements, an ACTOR-based generator, e.g., AvatarCLIP, needs to search in the latent space, which might be inefficient and unstable. Instead, our method can control the generation via motion prompt explicitly, which is more transparent and controllable.} During inference, the motion encoder is discarded and the latent feature can be randomly sampled from $\mathcal{N}(0; 1)$. Formally, the loss function $\mathcal{L}_{m2m}$ for the motion generator is
\begin{align}
 \notag \mathcal{L}_{rec}(p_t', p_t) &= ||p_t - p_t'||_2 + ||v_t - v_t'||_2\\
 \mathcal{L}_{m2m}(m', m) &= \sum_{p_t',p_t\in m', m} \mathcal{L}_{rec}(p_t',\  p_t) + \lambda_\text{KL}\mathcal{L}_\text{KL},
\end{align}
where $v = \mathcal{M}_\text{SMPL}(p)$ and $\mathcal{L}_\text{KL}$ is the KL-divergence regularization term to pull the predicted latent features to the normal distribution. 
After pretraining, the motion generator can be used to generate motion for downstream tasks by using the motion prompts, i.e., masked motions.

\subsection{Prompt Construction}
Since the mask in the motion prompt is provided by the motion generator, we only need to synthesize the unmasked poses to construct the motion prompt. In other words, we should use texts to synthesize the unmasked poses. To achieve this, our OOHMG learns a text-to-pose generator that takes in texts and predicts the poses. 
To provide supervision during training, we propose the first text-pose alignment model, TPA, based on the large-scale text-image alignment model CLIP. And to cover as diverse text as possible, we adopt an extreme yet effective training paradigm, i.e., wordless training. Below, we detail each procedure.

\begin{figure}[t]
	\centering
	\begin{center}
		\centering
		\includegraphics[clip=True, trim={0, 0, 0, 0}, width=\linewidth]{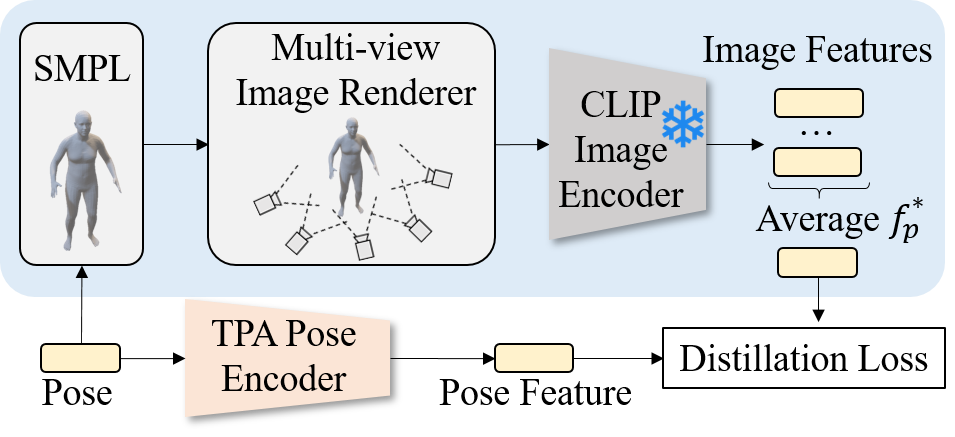}
	\end{center}
	\vspace{-7mm}
	\caption{The learning process of the TPA pose encoder. The upper part is the pipeline to extract the pose features via reusing the CLIP image encoder. TPA learns an end-to-end pose encoder that takes in the poses and predicts the output of the pipeline.}
	\vspace{-3mm}
	\label{fig:pipelinedistillation}
\end{figure}
\textbf{Text-pose alignment model.} 
Due to the lack of massive paired text-pose data, it is non-trivial to learn TPA from scratch. To this end, our TPA reuses the text encoder of CLIP. As for the pose encoder, TPA mines 3D pose knowledge of the CLIP image encoder. In fact, TPA is not the first work to leverage the CLIP image encoder for pose feature extraction. In AvatarCLIP~\cite{hong2022avatarclip}, the researchers extract pose features via the pipeline ``pose$\rightarrow$generate SMPL meshes$\rightarrow$render multi-view images$\rightarrow$use CLIP to extract image features$\rightarrow$average the features to obtain the final image features", as shown in the upper part of Fig.~\ref{fig:pipelinedistillation}. \jf{As testified in AvatarCLIP, this strategy should be enough for zero-shot text-pose alignment. And in our experiments, we also find it enough for our generator to learn to generate visually-plausible and text-consistent poses. Nevertheless, we do think it can be further improved by preserving the angle information using techniques like the view-dependent conditioning of DreamFusion~\cite{poole2022dreamfusion}} Unfortunately, AvatarCLIP~\cite{hong2022avatarclip} found this pipeline difficult to supervise the pose generation. \jf{In ~\cite{jain2022zero}, the researchers found that, if the generation space is too unconstrained, training solely with CLIP loss will result in severe artifacts that satisfy CLIP loss but with unrealistic geometry like Deep Dream artifacts \cite{olah2017feature}}. The potential reason is that CLIP has been trained on diverse images and there might be diverse solutions for the same text, causing the optimization divergent. To address this problem, our TPA limits the solution space to the 3D poses by distilling a tailored yet specific pose encoder. Specifically, we adopt an end-to-end pose encoder $E_p$ for mapping the poses $p$ to their features $f_p^*$ as shown in Fig.~\ref{fig:pipelinedistillation}. The distillation objective $\mathcal{L}_{E_p}$ is:
\begin{align}
   \mathcal{L}_{E_p}(p, f_p^*) = || E_p(p) - f_p^* ||_2 - \text{cossim}\big(E_p(p), f_p^*\big), \label{equ:pipelinedistillation}
\end{align}
where the first term of Equ.(\ref{equ:pipelinedistillation}) is for reducing the element-level distance between features. While the second term of Equ.(\ref{equ:pipelinedistillation}) is to reduce the angular difference.

\begin{figure}[t]
	\begin{center}
		\includegraphics[clip=True, trim={15 5 0 10}, width=\linewidth]{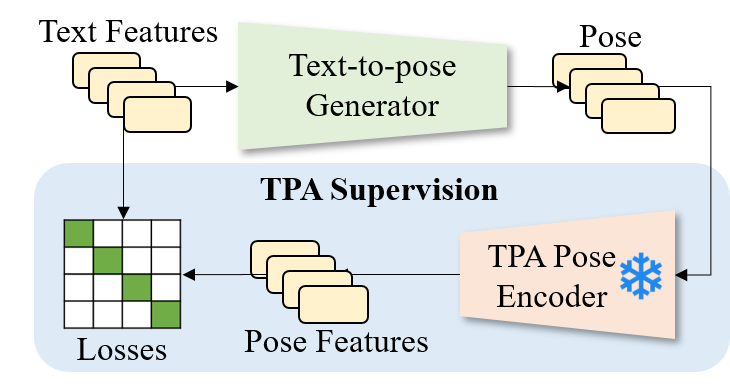}
	\end{center}
	\vspace{-6mm}
	\caption{Training process of text-to-pose generator. During training, the input features are randomly sampled from the feature space of the text encoder.}
	\vspace{-3mm}
	\label{fig:text2pose}
\end{figure}

\textbf{Wordless training for generalized text-to-pose generator.} To generate poses for open-vocabulary texts, the text-to-pose generator should train with as diverse texts as possible. However, as the text space is combinatorial, it's impractical to enumerate all possible texts for training. Nevertheless, since the texts will be encoded into text features by the text encoder of TPA to measure the alignment with the pose, it occurs to us that we can directly build a text-to-pose generator upon the normalized text feature space of TPA instead of real-world text space. By this means, it becomes trivial to obtain diverse inputs for training. To obtain diverse text features $f_d$, we sample them from Normal or Uniform distribution randomly by:
\begin{align}
    f_d = \frac{\epsilon+b}{|\epsilon+b|}, \ \ \ \forall \epsilon_i \in \epsilon, \epsilon_i \sim \mathcal{N}(0; 1) \text{\ or } \mathcal{U}[-1, 1],
\end{align} where $b \sim \mathcal{U}[-1, 1]$ is a random bias to avoid the features sampled around the zero features. We don't take the scale into account since there is a normalization operator.

Our text-to-pose generator $G_\text{t2p}$ takes a batch of sampled features $\mathbf{f_d}=[f_{d_1}, ..., f_{d_B}]$ as the inputs and predict the latent poses $\mathbf{p}^l = G_\text{t2p}(\mathbf{f_d})$ of VPoser. As mentioned in the preliminary, $\mathbf{p}^l$ can be decoded into the poses $\mathbf{p}$ by the decoder of VPoser. We can regulate the generator to predict in-distribution poses of VPoser by pulling the $\mathbf{p}^l$ close to the prior distribution of VPoser, i.e., the normal distribution. The optimization target is to minimize:
\begin{align}
\mathcal{L}_\text{t2p}(\mathbf{p}^l, \mathbf{p}, \mathbf{f_d}) = \mathcal{L}_\text{TPA}\Big(E_p(\mathbf{p}), \mathbf{f_d}\Big) + \lambda_\text{L2}||\mathbf{p}^l||_2, \label{equ:t2p}
\end{align}
where the second term of the loss function is used to regulate the predicted latent pose to close to the prior distribution of VPoser. And $\mathcal{L}_\text{TPA}$ is the same as $\mathcal{L}_\text{CLIP}$ in Equ.(\ref{equ:cliploss}) with image features replaced by pose features. The optimization process is illustrated in Fig.~\ref{fig:text2pose}.

\jf{\textbf{Overall Training Procedure.} Different modules of our method are trained separately since they have different training data. Specifically, we train the motion generator and TPA using AMASS data. After that, we train the text-to-pose generator with the frozen TPA and the wordless training strategy. During inference, a text is first encoded by the CLIP text encoder into text features. Then, the text-to-pose generator generates the text-consistent pose according to the text features, which are used for constructing the motion prompt. And the motion prompt will drive the motion generator to reconstruct the full motion.}

\section{Experiments}
We first introduce the datasets and baseline methods used in our experiments. Next, we evaluate the overall performance of zero-shot open-vocabulary human motion generation. And we also compare the performance of text-to-pose generation. After that, we conduct ablation studies to better understand our method.


\textbf{General Settings.} In our experiments, all motion data and textual descriptions originated from AMASS~\cite{NaureenMahmood2019AMASSAO} and BABEL~\cite{AbhinandaRPunnakkal2021BABELBA}, respectively. AMASS unifies various optical marker-based mocap datasets with more than 40 hours of motion data without textual labels. Following the same settings in ~\cite{hong2022avatarclip,MathisPetrovich2021ActionConditioned3H}, we down-sample the motion capture framerate to 30 per second and limit the duration of a motion to 2 seconds. As for the text data, BABEL is a dataset of textual sentences for motions. We remove lengthy sentences which exceed the CLIP's maximum text length of 77, resulting in a dataset with a size of 4178. In our paper, \jf{the checkpoint of CLIP (``CLIP-ViT-B/32") is used}. More about training details such as hyperparameters are presented in the supplementary. The code will be released.

\textbf{Baselines.}
In the following, we enumerate the related baseline methods. To the best of our knowledge, our work is the first offline zero-shot open-vocabulary text-to-motion generation. Therefore, we include a baseline of online zero-shot open-vocabulary text-to-motion generation, i.e., AvatarCLIP~\cite{hong2022avatarclip}, and a baseline of offline supervised open-vocabulary text-to-motion generation, i.e., MotionCLIP~\cite{tevet2022motionclip}. Similar to ours, AvatarCLIP also includes a text-to-pose phase via matching, and uses the matched poses to search the motion in the latent space of a pretrained motion VAE. Therefore, in evaluating text-to-pose generation, matching, as well as other baselines considered in AvatarCLIP~\cite{hong2022avatarclip}, are also included in our experiments. As for our OOHMG, except for the experiment for measuring controllability, we generate poses for texts and place the generated poses in the middle of the masked motions. 
More details about the baselines are placed in the supplementary.

\begin{figure*}[t]
	\begin{center}
		\includegraphics[clip=True, trim={0 40 40 10}, width=0.95\linewidth]{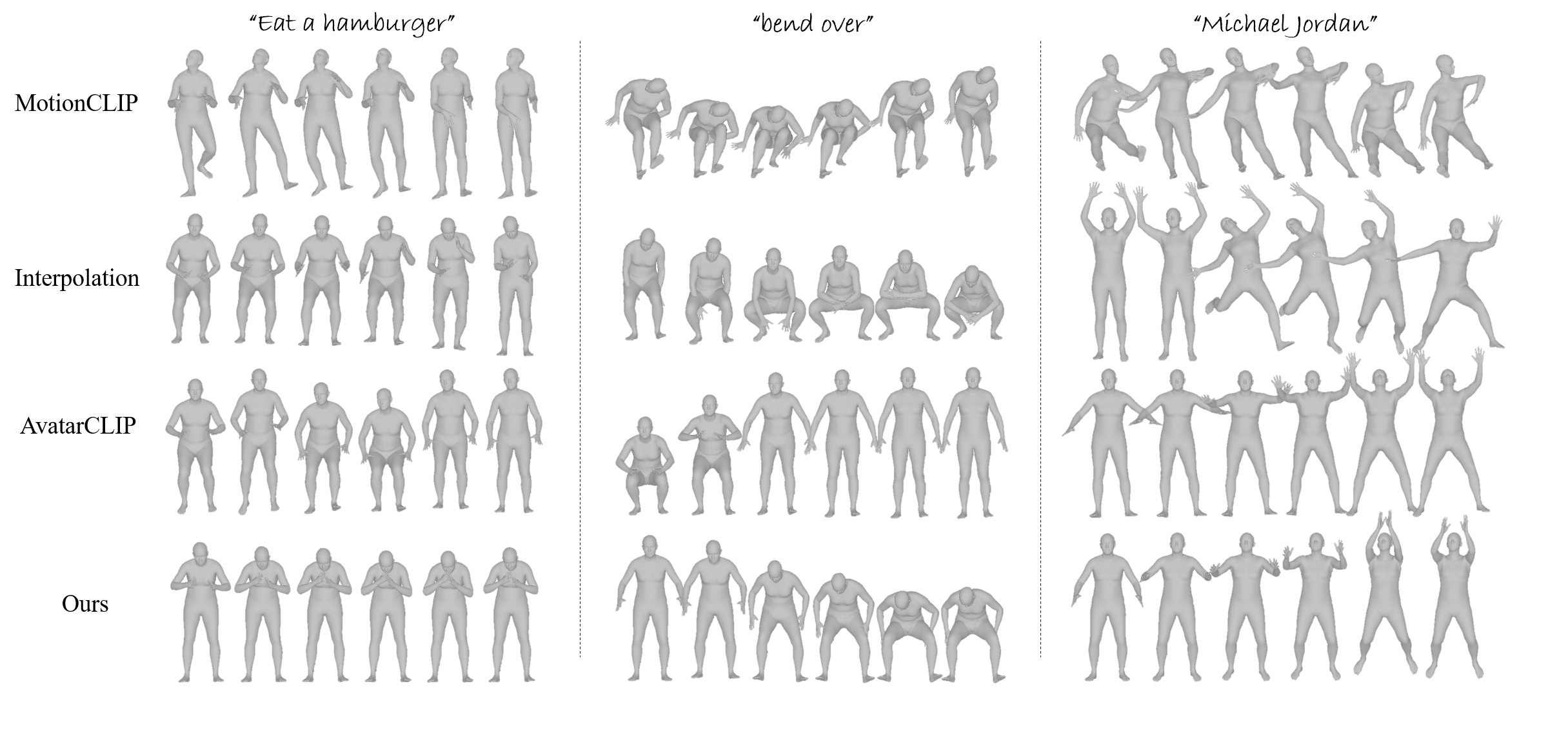}
	\end{center}
	\vspace{-7mm}
	\caption{The visual comparison results of different open-vocabulary text-to-motion methods. The results are part of the generated motions due to the space limit.}
	\vspace{-4mm}
	\label{fig:cmp_results}
\end{figure*}
\subsection{Open-vocabulary Text-to-Motion Generation}
In this part, we are interested in testifying about the ability to generate text-consistent motion across different baselines. For this purpose, we evaluate 1) whether the generated contents follow real-world motion dynamics, and 2) whether the generated motions are text-consistent.



\textbf{Motion Dynamics.} To answer the first question, we propose to measure the distance between the generated contents and the distribution of the real-world motion. Therefore, we train a general motion VAE upon AMASS, and use the average reconstruction error of the motion VAE to indicate the in-distribution degree (In-distrib.) of the baselines. Please refer to the supplementary for more details about the motion VAE. The larger the In-distrib. is, the more distant the generated contents are to the real-world motion distribution. We use \jf{BABEL} as textual descriptions to generate motions. As reported in Tab.~\ref{tab:text2motion}, our method outperforms the others by a clear margin. Particularly, we found that Interpolation~\cite{hong2022avatarclip} generates motion through linear interpolation without considering the motion dynamics, resulting in poor In-distrib. results. We also notice that the parameterized MotionCLIP performs poorly. The potential reason might be the gap of latent space between MotionCLIP and the CLIP text encoder. As shown in Fig.~\ref{fig:cmp_results}, we observe that MotionCLIP is likely to generate twisted poses while AvatarCLIP and ours are more natural. 

\begin{figure*}[t]
	\begin{center}
		\includegraphics[clip=True, trim={0 20 40 10}, width=0.9\linewidth]{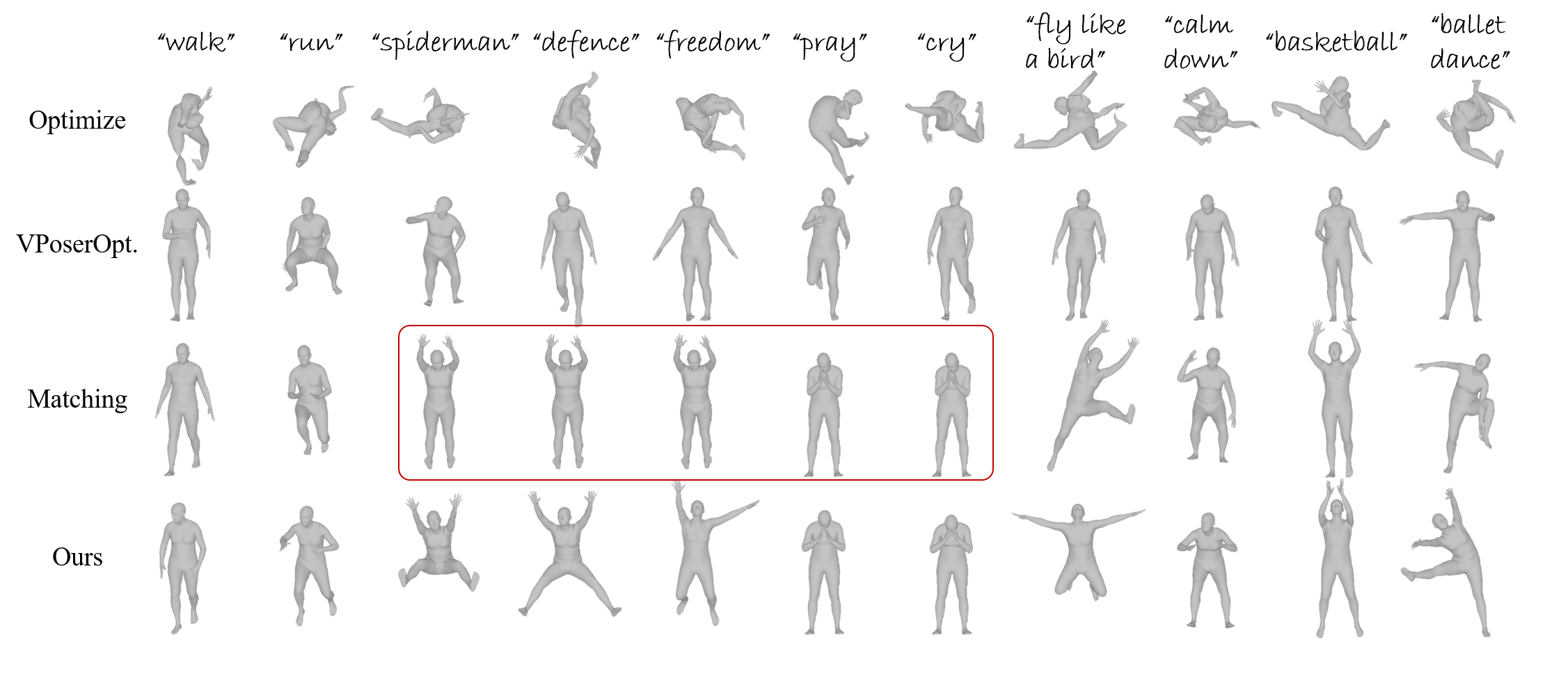}
	\end{center}
	\vspace{-7mm}
	\caption{The visual results of different text-to-pose generation methods. Besides ours, the other baseline methods require online matching or optimizations.}
	\vspace{-3mm}
	\label{fig:cmp_pose_results}
\end{figure*}
\begin{table}[t]
	\centering
	\caption{Comprehensive results for open-vocabulary text-to-motion generation. The arrow $\uparrow$ indicates the performance is better if the value is higher.}
 \vspace{-3mm}
        \setlength{\tabcolsep}{1.3mm}
	\begin{tabular}{c|c|ccc}
		\toprule
		 & In-distrib.$\downarrow$ & Top1$\uparrow$ & Top10$\uparrow$ & Top50$\uparrow$  \\ \midrule
		MotionCLIP~\cite{tevet2022motionclip}  &    0.2191    &  0.0029    &  0.0153     &   0.0661     \\
		Interpolation~\cite{hong2022avatarclip}  &  0.0312  & 0.0045   & 0.0472    &  0.1927   \\
		AvatarCLIP~\cite{hong2022avatarclip} &  0.0407  &    0.0002  &   0.0069    &  0.0290      \\ \midrule
  Ours       &  \textbf{0.0205}   &  \textbf{0.0792}   & \textbf{0.3231}    & \textbf{0.6494}    \\ \bottomrule  
	\end{tabular}
\vspace{-3mm}
	\label{tab:text2motion}
\end{table}
\textbf{Generating Text-consistent Motion.} Since both related baselines MotionCLIP and AvatarCLIP are CLIP-based, we use CLIP to measure the alignment between the motions and the texts. To this end, we extend the CLIP-R-precision~\cite{DongHukPark2021BenchmarkFC} to the level of motion to measure the text-motion alignment. Specifically, we say that the text-motion matching is accurate if, among all poses of the generated motion from different texts, the best-matched pose of the text is located in the generated motion of the text. To achieve a better motion-level CLIP-R-precision, the generated motion should \textbf{1) contain the text-consistent poses, 2) and does not contain irrelevant poses that might cause mismatching for other poses}. From the results at the right of Tab.~\ref{tab:text2motion}, we find that among all baselines, our method obtains the best TopK motion-level CLIP-R-Precision by a clear margin. It is worth noting that our method does not require online matching or optimization or paired text-motion training data like the baseline methods. We also observe that Interpolation~\cite{hong2022avatarclip} performs better than AvatarCLIP. One of the reasons is that, unlike Interpolation~\cite{hong2022avatarclip} which includes the condition poses as part of the generated motion, AvatarCLIP requires online optimization to obtain the motion, which is unstable and non-trivial to generate a motion consisting of the condition poses. And Interpolation~\cite{hong2022avatarclip} is less likely to generate new poses that might distract the matching process.

\begin{table*}[t]
	\centering
	\caption{Comparison among text-to-pose baselines. The arrow $\uparrow$ indicates the performance is better if the value is larger.}
 \vspace{-3mm}
	\begin{tabular}{cccc|ccc}
		\toprule
		& CLIP Score $\uparrow$         & In-distrib. $\downarrow$ & Cyc. Loss $\downarrow$ & Top1  $\uparrow$  & Top10  $\uparrow$  & Top50  $\uparrow$  \\ \midrule
		Matching~\cite{hong2022avatarclip}              & 0.2615 & \textbf{0.0015}  & 0.0288 & 0.0127 & 0.0831  & 0.2820         \\
		Optimize~\cite{hong2022avatarclip}       &  0.2455 &   0.8365  & 0.0047  & 0.0005 &	0.0038 &	0.0120   \\
		VPoserOptimize~\cite{hong2022avatarclip} &      0.2460  &  \textbf{0.0015}  & 0.0048 & 0.0005 & 0.0029 &	0.0168   \\ \midrule
		Ours      & \textbf{0.2694} & \textbf{0.0015}  & \textbf{0.0045}  & \textbf{0.0775} &	\textbf{0.3284} &	\textbf{0.6711} \\ \bottomrule
	\end{tabular}
	\label{tab:text2pose}
	\vspace{-3mm}
\end{table*}
\begin{table}[!t]
	\centering
	\caption{Controllability for pose-conditioned motion generation.}
 \vspace{-3mm}
	\begin{tabular}{c|ccc}
		\toprule
		& 1p$\downarrow$ & 2p$\downarrow$ & 3p$\downarrow$  \\ \midrule
		Interpolation~\cite{hong2022avatarclip}  & 0.0900  & 0.0865 & 0.0868   \\
		AvatarCLIP~\cite{hong2022avatarclip} & 0.8323 & 1.5982  & 2.1447  \\ \midrule
		Ours       & \textbf{0.0452}  &   \textbf{0.0131}   & \textbf{0.0129}  \\ \bottomrule  
	\end{tabular}
\vspace{-3mm}
	\label{tab:pose2motion}
\end{table}
\subsection{Prompt construction}
As described above, the prompt is in the form of masked motion. And OOHMG uses a text-to-pose generator to synthesize the unmasked poses of the masked motion according to the texts. Therefore, we are interested in 1) whether the generated motion can be controlled by the motion prompt, and 2) whether the generated poses are text-consistent.

\textbf{Controllability.} To answer question 1, we use the 4096 clustered poses used in AvatarCLIP~\cite{hong2022avatarclip} as the condition poses. We randomly sample from the clustered poses to construct the $\textit{K}\text{P}$ test set, where \textit{K} $\in \{1, 2, 3\}$ indicates the number of the unmasked poses of the motion prompt. We calculate the distance between $\textit{K}$ poses and the closest poses of the generated motions. As shown on the left of Tab.~\ref{tab:pose2motion}, we observe that our method also possesses the best controllability. Notice that, Interpolation~\cite{hong2022avatarclip} directly takes the condition poses as a part of the generated motion. However, there is still a small error for Interpolation~\cite{hong2022avatarclip} since interpolation is conducted on the latent codes of these poses in the latent space of VPoser and the encode-decode process causes the error. Differently, our method directly generates the motion and manages to obtain a smaller $\textit{K}\text{P}$ error. Nevertheless, we found that AvatarCLIP is difficult to generate motion that well-preserves the given poses. The potential reason is that AvatarCLIP requires optimizing the motion latent code in the high-dimensional latent space, which might be nonconvex and require a large number of optimization steps. 

\textbf{Open-vocabulary Text-to-Pose Generation.} 
To comprehensively understand our text-to-pose generator, we evaluate the generated poses from four aspects, i.e., the text-pose alignment (CLIP Score), the distance to the real-world pose distribution (In-distrib.), how much text information is preserved in the generated poses (Cycle loss for reconstructing text features from the generated poses) and CLIP-R-precision~\cite{DongHukPark2021BenchmarkFC} (TopK). For a detailed explanation of different metrics, \jf{please refer to the supplementary}. The baseline methods are adopted from AvatarCLIP since it is the only work that includes the zero-shot open-vocabulary text-to-pose generation, to our best knowledge. The results are represented in Tab.~\ref{tab:text2pose}. As we can see, the Matching method can obtain a higher CLIP score than Optimize and VPoserOptimize, which implies directly using CLIP to match is more effective than optimization via the complex pipeline as depicted in Fig~\ref{fig:pipelinedistillation}, which is also mentioned in AvatarCLIP. Nevertheless, Matching is unable to generate more accurate poses for diverse texts and therefore is less capable of preserving text information in the generated poses (i.e., high Cyc. loss in Tab.~\ref{tab:text2pose}). As shown in the red-circled region in Fig.~\ref{fig:cmp_pose_results}, Matching uses the same poses for texts with different meanings. By using TPA and wordless training, our text-to-pose generator obtains significant improvement across various metrics. Notice that, different from the baseline methods, our method does not require any online matching/optimization and does not see any real-world texts during the training phase. It means that our generator can be well-generalized to real-world texts.

\begin{table}[t]
	\centering
	\caption{Ablation results for our text-to-pose generator.}
 \vspace{-3mm}
        \setlength{\tabcolsep}{0.7mm}
	\begin{tabular}{cccc}
		\toprule
		& CLIP Score $\uparrow$         & In-distrib. $\downarrow$  & Top50  $\uparrow$  \\ \midrule
		VPoserOptimize~\cite{hong2022avatarclip} &      0.2460  &  \textbf{0.0015}   & 	0.0168   \\ \midrule
		Ours (Text+Score)  & 0.2620 & 0.1127 & 0.2090 \\
		Ours (Text+$\mathcal{L}_\text{TPA}$)  & 0.2601 & 0.1210 & 0.2104   \\
		Ours (Random+$\mathcal{L}_\text{TPA}$)  & \textbf{0.2711} & 0.0111  & \textbf{0.7224}    \\ \midrule
		Ours (Random+$\mathcal{L}_\text{t2p}$)   & 0.2694 & \textbf{0.0015}	& 0.6711    \\ \bottomrule
	\end{tabular}
	\label{tab:text2pose_abl}
	\vspace{-3mm}
\end{table}
\begin{figure}[t]
	\begin{center}
		\includegraphics[clip=True, trim={0 0 0 0}, width=\linewidth]{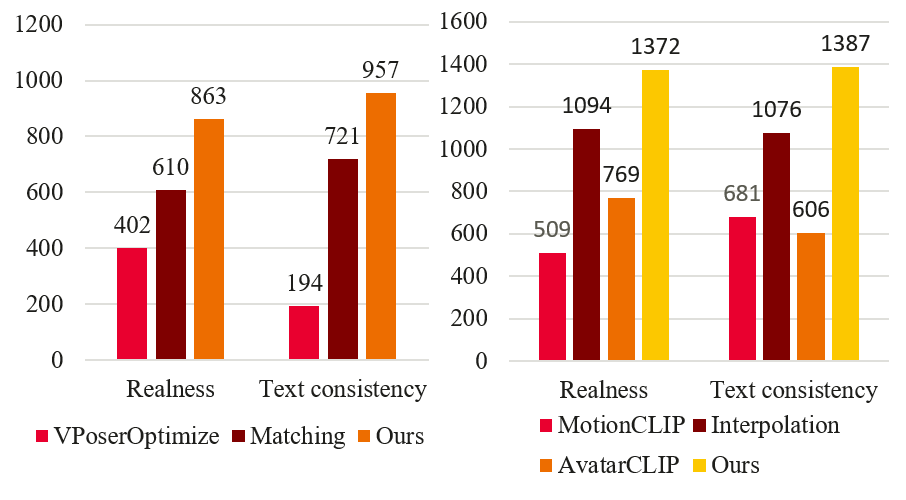}
	\end{center}
	\vspace{-7mm}
	\caption{Human evaluation for text-to-pose generation (Left) and text-to-motion generation (Right).}
	\vspace{-4mm}
	\label{fig:he_all}
\end{figure}
\textbf{Ablation studies.} As we can see in Tab.~\ref{tab:text2pose_abl}, by simply replacing the original pipeline with TPA, the text-to-pose generator that optimizes to maximize the TPA score, i.e., Ours (Text+Score), can significantly improve the CLIP score in comparison with VPoserOptimize. However, during our experiment, we observed that maximizing the TPA score as the objective is sensitive to the performance of TPA. And the stability of optimization can be further improved when minimizing the $\mathcal{L}_\text{TPA}$ instead (analysis in the supplementary). We contribute such stability to more dense supervision by drawing other samples into the contrastive loss. To this end, we suggest using $\mathcal{L}_\text{TPA}$ for the other experiments. Another valuable observation is that, by wordless training with randomly sampled text features, Ours (Random+$\mathcal{L}_\text{TPA}$) not only achieves the best CLIP score but also obtains significant improvement on the In-distrib. metric. The potential reason could be the infinite amount and diversity of training data. With a limited amount of training data, it's easier for the generator to exploit the difference between TPA and the original pipeline to obtain an overfit solution (e.g., generating strange or twisted poses) for the training data. Another evidence is that, by using wordless training, Ours (Random+$\mathcal{L}_\text{TPA}$) has better performance than Ours (Text+$\mathcal{L}_\text{TPA}$). 
It implies that using real texts for training might encourage the generator to overfit TPA, resulting in a poor CLIP score. 
To generate in-distribution poses, Ours (Random+$\mathcal{L}_\text{t2p}$) also includes the L2-norm regularization term which pulls the predicted latent pose to the center of the prior distribution of VPoser.

\subsection{Qualitative Results}
\textbf{Human Evaluation.} As for the qualitative results, We also conducted a series of human evaluations. We design a questionnaire that includes 50 queries for comparing different methods (25 for text-to-pose generation and 25 for text-to-motion generation). In each query, the participant was required to rank the performance of different methods in terms of realness and text consistency. For pose generation, we assign scores 2, 1, and 0 for the methods with ranks 1st, 2nd, and 3rd, respectively. And for motion generation, we assign scores 3, 2, 1, and 0 for the methods ranking 1st, 2nd, 3rd, and 4th, respectively. By the end of the submission, we have collected 25 available feedbacks and the total scores for each method are calculated and reported in Fig.~\ref{fig:he_all}. For more details, please refer to the supplementary. From the results, we observe that our methods for both pose and motion generations have obtained the best results. And we also find that these results are mostly in line with the quantitative results in previous experiments. It suggests that future works can follow the same evaluation protocol for this task.



\textbf{Efficiency.} Here, we also compare the maximum samples each method can handle simultaneously (i.e., batch size) and the time cost. \jf{The reported time is per batch and the batch size is 1, and is averaged over 100 repeated experiments.} The experiments are conducted using one NVIDIA V100 Tensor Core (32G). As shown in Tab.~\ref{tab:efficiency}, both our TPA and OOHMG can directly infer the results with significant improvements. As the results suggest, our OOHMG is the first real-time zero-shot text-to-motion generation method.

\begin{table}[t]
	\centering
	\caption{The inference efficiency of different methods. }
 \vspace{-3mm}
	\begin{tabular}{ccc}
		\toprule
             & Batch size  $\uparrow$   & Time (sec)  $\downarrow$  \\ \midrule
             Pipeline with CLIP~\cite{hong2022avatarclip} & 15 &  1.2068  \\
             Our TPA & $\sim$ \textbf{130K} & \textbf{0.0172} \\ \hline
             MotionCLIP~\cite{tevet2022motionclip} & 375 & 0.0242\\
             AvatarCLIP~\cite{hong2022avatarclip} & 9 & 140 \\
             Our OOHMG & $\sim$\textbf{14K} & \textbf{0.0159} \\
             \bottomrule
	\end{tabular}
	\label{tab:efficiency}
\end{table}

\section{Conclusion}
In this paper, we propose an offline open-vocabulary human motion generation (OOHMG) framework in a zero-shot learning manner, which draws inspiration from prompt learning. To address the difficulty of optimization with the complex pipeline, we propose the first text-pose alignment model which is efficient and effective for supervising the training of the pose generator. To handle diverse and unseen real-world texts, we propose a novel wordless training mechanism. Extensive experiments show that our method can generate better text-consistent poses and motions across various baselines and metrics.

\section*{Acknowledgments}
This work was supported in part by the National Key R\&D Program of China under Grant No.2021ZD0111601, in part by the Guangdong Basic and Applied Basic Research Foundation (NO.~2020B1515020048), in part by the National Natural Science Foundation of China (NO.~61976250), in part by the Shenzhen Science and Technology Program (NO.~JCYJ20220530141211024) and in part by the Fundamental Research Funds for the Central Universities under Grant 22lgqb25.

\appendix

\section{Model Structure and Training Details}
Our OOHMG consists of two generators, i.e., the text-to-pose and motion generators. To optimize the text-to-pose generator, we also distill a text-pose alignment model, namely TPA, from the versatile CLIP~\cite{AlecRadford2021LearningTV}. To this end, these three neural networks contribute to our OOHMG in this paper. In this part, we describe the format of input and output as well as the architecture for these networks.

\textbf{Fundamental neural network architectures.} There are mainly two kinds of neural network architectures used in this paper, i.e., ResNet-based networks and Transformer-based networks. For \textbf{ResNet-based networks}, input poses are projected into embeddings by a linear layer, and then processed by 6 residual blocks. The intermediate results are normalized by a layer normalization layer and another linear layer to obtain the final results. The residual block will first normalize the input by a layer normalization, and then forward the normalized embeddings to a Linear-GELU-Dropout(0.1)-Linear-Dropout(0.1) networks to predict the residual which will be added to the normalized embeddings to form the output of the residual block. The hidden size is 1024. As for \textbf{Transformer-based networks}, we adopt a similar architecture as Bert~\cite{devlin2018bert}. The architectures of the transformer encoder and decoder layer are implemented by PyTorch~\cite{AdamPaszke2019PyTorchAI}. The poses of a motion are first projected by a linear projection layer, then processed by an 8-layered transformer encoder/decoder, and finally fed to an estimation layer to obtain a prediction. The number of attention heads is 8, the hidden size is 1024 and the dropout rate is 0.1.

\begin{figure*}[t]
	\begin{center}
		\centering
		\includegraphics[clip, trim={0 0 0 0}, width=0.47\linewidth]{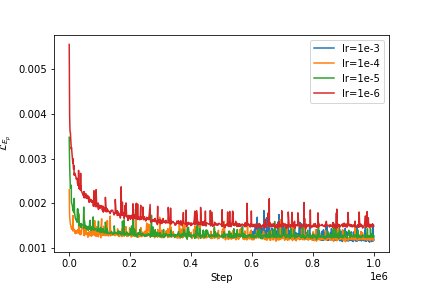}
  \includegraphics[clip, trim={0 0 0 0}, width=0.47\linewidth]{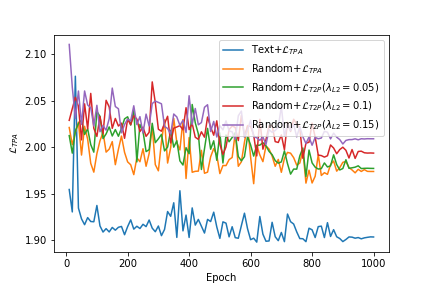}
		\caption{Training curves of our TPA and text-to-pose generator. The left figure includes the loss curves of TPA with different learning rates. The right figure includes the $\mathcal{L}_\text{TPA}$ curves of the text-to-pose generator with different training loss functions.}
		\label{fig:hyper_pose}
		\vspace{-3mm}
	\end{center}
\end{figure*}
\begin{figure*}[t]
	\begin{center}
		\centering
  \includegraphics[clip, trim={0 0 0 0}, width=0.47\linewidth]{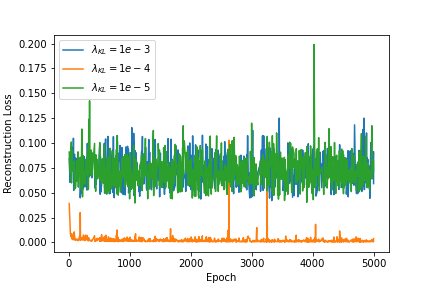}
  \includegraphics[clip, trim={0 0 0 0}, width=0.47\linewidth]{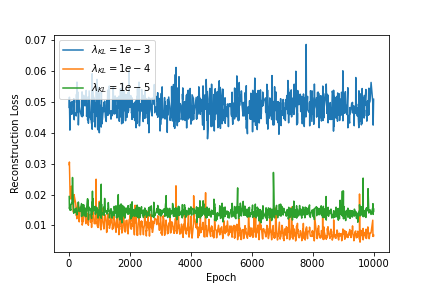}
		\caption{Training curves of our motion generator and general motion VAE which is used for evaluation. The left figure ablates the $\lambda_\text{KL}$ of our motion generator; The right figure ablates the $\lambda_\text{KL}$ of the motion VAE which is used for evaluation.}
		\label{fig:hyper_motion}
	\end{center}
\end{figure*}
\textbf{TPA.} TPA is distilled from CLIP for aligning 3D poses and texts. Specifically, for the text encoder, TPA simply reuses the text encoder of CLIP. As for the pose encoder, TPA adopts the ResNet-based network. TPA pose encoder takes in the 6D-rotation representation of the pose and predicts the output of the original pipeline. The batch size is 1024, and the learning rate is 1e-4 at the beginning and annealed by the CosineAnnealingLR scheduler implemented by PyTorch. The number of training iterations is 1e6. The training curves of loss in learning rates are plotted at the left of Fig.~\ref{fig:hyper_pose}. As for the process of the \textbf{original pipeline}, we mostly adopt the process used in AvatarCLIP~\cite{hong2022avatarclip}. Specifically, as shown in our manuscript, the 3D pose representation is first used to generate the 3D meshes by SMPL~\cite{SMPL-X:2019,LeonidPishchulin2017BuildingSS}. Then, 5 look-at cameras, with azimuth angles $[120, 150, 180, 210, 240]$ and fixed elevation, render the mesh into 5 images. After that, the image encoder of CLIP extracts the features of images and the pipeline takes the average for the features as the features of the 3D pose. The training poses are sampled from AMASS~\cite{NaureenMahmood2019AMASSAO}.

\begin{table}[t]
	\centering
	\caption{Ablation results for our text-to-pose generator with different $\lambda_\text{L2}$ evaluated on BABEL~\cite{AbhinandaRPunnakkal2021BABELBA}.}
	\begin{tabular}{cccc}
		\toprule
		& CLIP Score $\uparrow$         & In-distrib. $\downarrow$  & Top50  $\uparrow$  \\ \midrule
		$\lambda_\text{L2} = 0$  & \textbf{0.2711} & 0.0111  & \textbf{0.7224}    \\
	$\lambda_\text{L2} = 0.05$   & 0.2702 & 0.0019	& 0.7039    \\
	$\lambda_\text{L2} = 0.1$   & 0.2694 & {0.0015}	& 0.6711    \\
  $\lambda_\text{L2} = 0.15$   & 0.2689 & \textbf{0.0012}	& 0.6446    \\\bottomrule
		
	\end{tabular}
	\label{tab:l2norm}
\end{table}
\textbf{Text-to-pose generator.} The architecture of our text-to-pose generator is the ResNet-based network. It takes the text features extracted by TPA/CLIP text encoder and predicts the latent pose of the VPoser which is decoded by the VPoser decoder to obtain the 6D-rotation pose representation. During training, the 6D-rotation representation is fed to the TPA pose encoder for the pose feature. The batch size is 1024, and the learning rate is 1e-3 at the beginning and annealed by the CosineAnnealingLR scheduler implemented by PyTorch. The number of training epochs is 1K and the number of iterations of each epoch is 1K. As for selection for $\lambda_\text{L2}$, we found that when $\lambda_\text{L2}$ equals 0.1, the performances of different metrics are more in balance as shown in Tab.~\ref{tab:l2norm}. The loss curves with different $\lambda_\text{L2}$ are plotted in the right of Fig.~\ref{fig:hyper_pose}. As for noise features, the noise features are randomly sampled either from Normal distribution $\mathcal{N}(0; 1)$ or from Uniform distribution $\mathcal{U}[-1, 1]$. The proportions of the features from these two distributions are 50\% and 50\%. In addition, a random bias sampled from $\mathcal{U}[-1, 1]$ is added to each of the noise features. 

\textbf{Pretrained motion generator.} As described in the main text, our text-to-motion generation uses the combination of a pretrained motion model and pose prompt. And the pretrained motion model is the only network in this stage. The pretrained motion model uses the transformer-based network architecture. As described in our manuscript, the pretrained motion model includes a motion encoder and a motion generator. During the training phase, the motion encoder takes in a motion with 6D-rotation representations and two tokens for mean and standard deviation. The predicted mean and standard of the encoder are used to sample latent code via the reparameterization trick. The motion generator takes the latent code and randomly masks motion as input to reconstruct the complete motion. The batch size is 64, and the learning rate is 1e-4 at the beginning and annealed by the CosineAnnealingLR scheduler implemented by PyTorch. The number of training epochs is 5K. $\lambda_\text{KL}$ is set as 1e-4 empirically. The reconstruction loss for different $\lambda_\text{KL}$ is shown at the left of Fig.~\ref{fig:hyper_motion}. During inference, the values of the latent code are set as 0.

\begin{figure*}[t]
	\begin{center}
		\includegraphics[clip=True, trim={0 20 40 10}, width=0.95\linewidth]{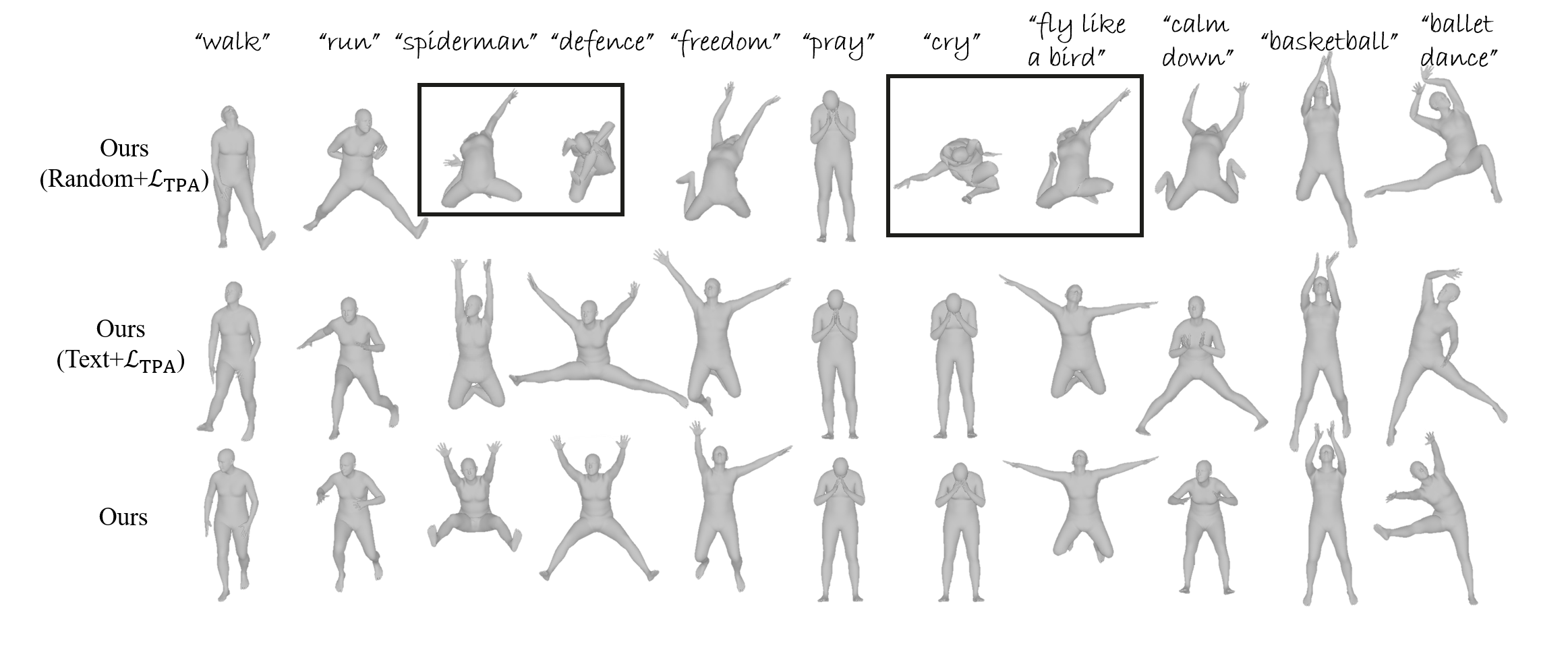}
	\end{center}
	\vspace{-7mm}
	\caption{\jf{The visual results of different combinations of wordless training and real-world text supervision.}}
	\label{fig:wordless_abl}
\end{figure*}
\section{Experiment Details}
\subsection{Baselines Details.} The results of all baseline methods are obtained by running their open-released codes. As for MotionCLIP~\cite{tevet2022motionclip}, we directly adopt their open-released model for motion generation. As for Interpolation / Matching / Optimize / VPoserOptimize / AvatarCLIP~\cite{hong2022avatarclip}, we adopt and use their open-released code and make small revisions for evaluations. Particularly, Interpolation and AvatarCLIP are originally developed for generating motion using Top5 poses of a text. Therefore, these methods are designed to use a fixed number of condition poses with similar semantics. In our experiments, to evaluate the controllability, the semantics and number of condition poses are different. Thus, we adapt the original code to evaluate the controllability. In the other experiment we simply their original code for evaluation.

\begin{table}[t]
    \centering
    \caption{CLIP scores of our text-to-pose generator trained with maximizing TPA score or $\mathcal{L}_\text{TPA}$ using the checkpoints of TPA at iteration 1e4, 1e5 and 1e6.}
    \begin{tabular}{c|ccc}
    \toprule
        Iterations & 1e4 & 1e5 & 1e6 \\ \hline
        Text+Score & 0.2491 & 0.2524 & 0.2620 \\
        Text+$\mathcal{L}_\text{TPA}$ & 0.2557 & 0.2613 & 0.2601 \\ \bottomrule
    \end{tabular}
    \label{tab:score_vs_cont}
\end{table}

\begin{table}[t]
	\centering
	\caption{Comprehensive results with/without adding an initial pose to the prompt. The arrow $\uparrow$ indicates the performance is better if the value is higher.}
 \vspace{-3mm}
        \setlength{\tabcolsep}{1.1mm}
	\begin{tabular}{c|c|ccc}
		\toprule
		 & In-distrib.$\downarrow$ & Top1$\uparrow$ & Top10$\uparrow$ & Top50$\uparrow$  \\ \midrule
		Ours wo init. pose       &  {0.0208}   &  {0.0768}   & {0.3135}    &  {0.6154}    \\ \bottomrule  
  Ours       &  \textbf{0.0205}   &  \textbf{0.0792}   & \textbf{0.3231}    & \textbf{0.6494}    \\ \bottomrule  
	\end{tabular}
	\label{tab:prompt}
\end{table}
\subsection{OOHMG Prompt Details}
Given a text, our text-to-pose generator synthesizes the text-consistent pose and places it in the middle of a sequence of masks to form the prompt. However, we found that in this manner, the motion generator usually generates a motion filled with similar poses. This is reasonable since the motion filled with similar poses might also exist in the real world. Thanks to the strong controllability of the motion generator, \jf{we can easily adjust the generated motion by refining the prompts with multiple poses. For example, we can use descriptions to specify the fore-pose, middle pose, and post-pose to construct a motion prompt. In our experiments,} we found that adding an initial pose (which latent poses of VPoser is a zero vector) to the prompt can significantly improve the variation of the generated motions. And we also found that in this manner, the performance of motion evaluation also improve a little bit, as presented in Tab.~\ref{tab:prompt}.

\subsection{Wordless Training Visualization}
\jf{The poses in Fig.~\ref{fig:wordless_abl} are generated w.o. / w. wordless training, corresponding to ``Ours(Text+$\mathcal{L}_\text{TPA}$) / Ours(Random+$\mathcal{L}_\text{TPA}$)" in Tab.4 of the main paper. We observe that, without wordless training, the generator may not perform well with some unseen texts (in boxes).}

\subsection{\jf{Visualization for poses generated with random text features}}
\begin{figure}[t]
    \centering
    \includegraphics[width=\linewidth, clip=true, trim={0pt 60pt 10pt 30pt}]{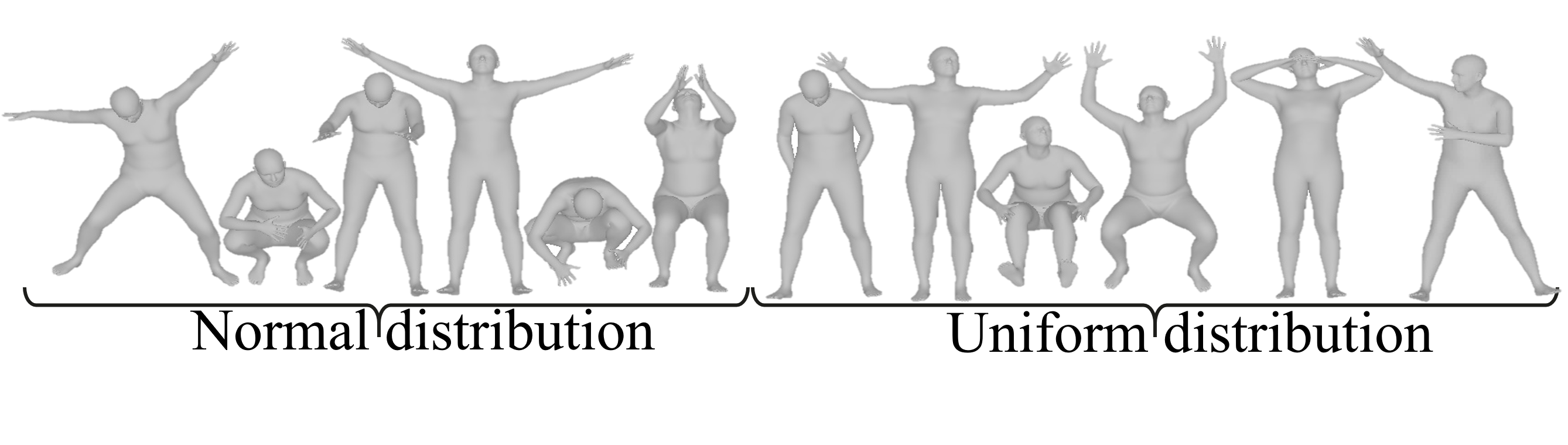}
    \caption{\jf{The visualization of poses generated with random text features.}}
    \label{fig:pose_random_text}
\end{figure}

\jf{In Fig.~\ref{fig:pose_random_text}, we show several poses generated by our text-to-pose generator using random features obeying Equ.(5) in the main paper. The results imply that our method can generalize to random texts.}

\subsection{Contrastive Loss and Maximizing Score}

As we can see in Tab.~\ref{tab:score_vs_cont}, when TPA is finished training at iter 1e6, by maximizing the TPA score (i.e., Text+Score) can also obtain comparable performance to Text+$\mathcal{L}_\text{TPA}$ which uses $\mathcal{L}_\text{TPA}$ for optimization. However, if TPA is not converged, Text+Score is more likely to have a degenerated performance. While Text++$\mathcal{L}_\text{TPA}$ has more stable performance. We contribute such stability to more dense supervision by drawing other samples into the contrastive loss.

\subsection{Evaluation Metrics.} In our experiments, we mainly evaluate text-to-pose and text-to-motion generations. Unfortunately, there are no suitable evaluation metrics in the current literature. In AvatarCLIP~\cite{hong2022avatarclip}, they only conduct user studies. Therefore, our paper proposes to adopt and adapt popular metrics to evaluate performance.

As for text-to-pose generation, we mainly evaluate the CLIP similarity score (i.e., \textbf{CLIP Score}), in-distribution distance (i.e., \textbf{In-distrib.}), text-to-pose-to-text reconstruction loss (i.e. \textbf{Cycle loss}) and  CLIP-R-precision~\cite{DongHukPark2021BenchmarkFC} (i.e., \textbf{TopK}). Specifically, In-distrib. is the reconstruction loss of the VPoser~\cite{SMPL-X:2019}. If the generated pose is similar to the training poses of VPoser, the reconstruction is likely to be small. And for Cyc. Loss, we train an auxiliary neural network for each method to learn the reverse mapping from the generated poses to their corresponding text features. And if the regression loss is smaller, the generated poses are more likely to carry more textual information and thus more diverse. The structure of the auxiliary regression model is similar to the pose encoder of TPA. The hidden size is 512 and the number of ResBlock is 2, and no dropout. The learning rate is 1e-3 and the number of iterations is 1e4.

As for text-to-motion generation, we mainly evaluate the in-distribution degree (i.e., \textbf{In-distrib.}) and the extended CLIP-R-precision for motion (i.e., \textbf{TopK}). Specifically, In-distrib. also uses reconstruction loss of a pretrained motion VAE. The pretrained motion VAE is similar to the architecture of ACTOR~\cite{MathisPetrovich2021ActionConditioned3H} without condition input. The KL loss term is 1e-4 as ablated in the right of Fig.~\ref{fig:hyper_motion}. As for the extended CLIP-R-precision, we say that the text-motion matching is accurate if, among all poses of the generated motion from different texts, the best-matched pose of the text is located in the generated motion of the text. To achieve a better motion-level CLIP-R-precision, the generated motion should 1) contain the text-consistent poses, and 2) and does not contain irrelevant poses that might cause mismatching for other poses.

\jf{To measure whether the motion generator can synthesize motion according to the given poses, we also introduce the $K$P metric. We randomly sample poses from 4096 clustered poses from AMASS and use them as conditional poses. And we measure whether the generated motion contains that pose by calculating the minimal reconstruction error of these poses. The small the reconstruction error is, the better the generated motion preserves the conditional poses. We use the 4096 clustered poses used in AvatarCLIP~\cite{hong2022avatarclip} as the condition poses. We randomly sample from the clustered poses to construct the $\textit{K}\text{P}$ test set, where \textit{K} $\in \{1, 2, 3\}$ indicates the number of the condition poses used for generation a motion. The measurement of $\textit{K}$P for a generated motion $m$ conditioned on poses $\{p'_k\}_{j=1:K}$ is formulated as:
\begin{align}
	\textit{K}\text{P}(m, \{p'_k\}_{j=1:K}) = \frac{1}{K} \sum_{k=1:K} \min_{p_j \in m} || p'_k - p_j ||_2.
 \label{equ:kp}
\end{align}}

\subsection{Human Evaluation}
For human evaluation, \jf{we designed our human evaluation questionnaires in the free online platform (https://wj.qq.com/). We shared the questionnaires on the internet with the non-paid and unknown subjects who are not participated in our work, including but not limited to colleagues from different universities, workers from different industrial companies, etc. For each question in the questionnaire, a subject will be provided with a text description and several shuffled generation contents from different methods, following two queries in terms of text consistency and realness.} We randomly invite 25 human evaluators to compare the performance of pose generation and motion generation of different methods. For each participant, we inquire about 50 questions (25 for pose and 25 for motion). A text and the generated poses/motions of different methods are given for each question. The participants are required to give the order of methods in terms of realness and text consistency. For realness, we ask the participant which pose/motion is more vivid as real-world pose/motion. And for text consistency, we ask them which pose/motion is more in line with the given text. To avoid the participant trivially giving the meaningless order, we randomly change the order of the presentation order of different methods. There are three pose generation methods and four motion generation methods. For the pose generation method with ranks 1st, 2nd, and 3rd, we assign scores 3, 2 ad 1 for each question. Similarly, for the motion generation method with ranks 1st, 2nd, 3rd, and 4th, we assign scores 4, 3, 2 ad 1 for each question. To better understand the content of the human evaluation, we also include the visualization of poses and motions of different methods, which are placed in a separate folder along with the supplementary.


\subsection{Discussion and Limitations}
As foundation models, e.g., CLIP, become more mature and learn more real-world knowledge, it provides us with new opportunities and challenges to a new learning paradigm. In this paper, we show one of the possibilities that learning from the foundation model instead of learning from data. We believe such attempts have an advantage over learning from data since the foundation model can better associate multi-modality data to make better decisions. Particularly, in our method, we found that using noisy training data can probe diverse knowledge out of the foundation model, which implies the feasibility of building an agent that can actively and continuously learn knowledge from the foundation model starting from chaos, i.e., noises, without manually feeding data which might limit the learnable knowledge of the foundation model. By this means, the agent might be able to learn something that is existed but we have not thought of yet or tasks we cannot formulate mathematically using our current knowledge. 

\jf{Although our method is mainly offline generation, our method can also be extended to online generation. In addition to pure online generation, ours may provide a better initial solution to speed up the optimization and improve robustness.}


However, as one of the few pioneers, several aspects can be improved in our work. One is that CLIP learns from static image data and cannot handle motion description. It cannot handle some difficult texts like a sentence having multiple successive motions. However, this can be addressed by a divide-and-conquer strategy. And with the great controllability of our method, our methods can be easily extended to handle this problem. \jf{And our work mainly evaluates the existing methods using CLIP-based measurements, e.g., CLIP-R-precision, since the compared methods are mostly CLIP-based.} Nevertheless, there are several foundation models for aligning video and texts, but we found that most of them are learning with limited types of video data and are not as general as CLIP due to the difficulty of data collection for video training data. To this end, in our paper, we still prefer CLIP for zero-shot learning. And we leave the research with other foundation models in the future.

{\small
\bibliographystyle{ieee_fullname}
\bibliography{camera_arxiv}

\begin{thebibliography}{10}\itemsep=-1pt

\bibitem{GunjanAggarwal2021Dance2MusicAD}
Gunjan Aggarwal and Devi Parikh.
\newblock Dance2music: Automatic dance-driven music generation.
\newblock {\em arXiv: Sound}, 2021.

\bibitem{HyeminAhn2018Text2ActionGA}
Hyemin Ahn, Timothy Ha, Yunho Choi, Hwiyeon Yoo, and Songhwai Oh.
\newblock Text2action: Generative adversarial synthesis from language to
  action.
\newblock {\em international conference on robotics and automation}, 2018.

\bibitem{ChaitanyaAhuja2019Language2PoseNL}
Chaitanya Ahuja and Louis-Philippe Morency.
\newblock Language2pose: Natural language grounded pose forecasting.
\newblock {\em international conference on 3d vision}, 2019.

\bibitem{TEACH:3DV:2022}
Nikos Athanasiou, Mathis Petrovich, Michael~J. Black, and G{\"u}ll Varol.
\newblock {TEACH}: {T}emporal {A}ction {C}ompositions for {3D} {H}umans.
\newblock In {\em {International Conference on 3D Vision (3DV)}}, 2022.

\bibitem{brown2020language}
Tom Brown, Benjamin Mann, Nick Ryder, Melanie Subbiah, Jared~D Kaplan, Prafulla
  Dhariwal, Arvind Neelakantan, Pranav Shyam, Girish Sastry, Amanda Askell,
  et~al.
\newblock Language models are few-shot learners.
\newblock {\em Advances in neural information processing systems},
  33:1877--1901, 2020.

\bibitem{cai2022humman}
Zhongang Cai, Daxuan Ren, Ailing Zeng, Zhengyu Lin, Tao Yu, Wenjia Wang,
  Xiangyu Fan, Yang Gao, Yifan Yu, Liang Pan, et~al.
\newblock Humman: Multi-modal 4d human dataset for versatile sensing and
  modeling.
\newblock {\em arXiv preprint arXiv:2204.13686}, 2022.

\bibitem{cai2021playing}
Zhongang Cai, Mingyuan Zhang, Jiawei Ren, Chen Wei, Daxuan Ren, Jiatong Li,
  Zhengyu Lin, Haiyu Zhao, Shuai Yi, Lei Yang, et~al.
\newblock Playing for 3d human recovery.
\newblock {\em arXiv preprint arXiv:2110.07588}, 2021.

\bibitem{devlin2018bert}
Jacob Devlin, Ming-Wei Chang, Kenton Lee, and Kristina Toutanova.
\newblock Bert: Pre-training of deep bidirectional transformers for language
  understanding.
\newblock {\em arXiv preprint arXiv:1810.04805}, 2018.

\bibitem{KevinFrans2021CLIPDrawET}
Kevin Frans, L.~B. Soros, and Olaf Witkowski.
\newblock Clipdraw: Exploring text-to-drawing synthesis through language-image
  encoders.
\newblock {\em arXiv: Computer Vision and Pattern Recognition}, 2021.

\bibitem{Guo_2022_CVPR}
Chuan Guo, Shihao Zou, Xinxin Zuo, Sen Wang, Wei Ji, Xingyu Li, and Li Cheng.
\newblock Generating diverse and natural 3d human motions from text.
\newblock In {\em Proceedings of the IEEE/CVF Conference on Computer Vision and
  Pattern Recognition (CVPR)}, pages 5152--5161, June 2022.

\bibitem{ChuanGuo2020Action2MotionCG}
Chuan Guo, Xinxin Zuo, Sen Wang, Shihao Zou, Qingyao Sun, Annan Deng, Minglun
  Gong, and Li Cheng.
\newblock Action2motion: Conditioned generation of 3d human motions.
\newblock {\em acm multimedia}, 2020.

\bibitem{ho2022imagen}
Jonathan Ho, William Chan, Chitwan Saharia, Jay Whang, Ruiqi Gao, Alexey
  Gritsenko, Diederik~P Kingma, Ben Poole, Mohammad Norouzi, David~J Fleet,
  et~al.
\newblock Imagen video: High definition video generation with diffusion models.
\newblock {\em arXiv preprint arXiv:2210.02303}, 2022.

\bibitem{hong2022avatarclip}
Fangzhou Hong, Mingyuan Zhang, Liang Pan, Zhongang Cai, Lei Yang, and Ziwei
  Liu.
\newblock Avatarclip: Zero-shot text-driven generation and animation of 3d
  avatars.
\newblock {\em arXiv preprint arXiv:2205.08535}, 2022.

\bibitem{CatalinIonescu2014Human36MLS}
Catalin Ionescu, Dragos Papava, Vlad Olaru, and Cristian Sminchisescu.
\newblock Human3.6m: Large scale datasets and predictive methods for 3d human
  sensing in natural environments.
\newblock {\em IEEE Transactions on Pattern Analysis and Machine Intelligence},
  2014.

\bibitem{jain2022zero}
Ajay Jain, Ben Mildenhall, Jonathan~T Barron, Pieter Abbeel, and Ben Poole.
\newblock Zero-shot text-guided object generation with dream fields.
\newblock In {\em CVPR}, pages 867--876, 2022.

\bibitem{NikolayJetchev2022ClipMatrixTC}
Nikolay Jetchev.
\newblock Clipmatrix: Text-controlled creation of 3d textured meshes.
\newblock 2022.

\bibitem{lin2018human}
Xiao Lin and Mohamed~R Amer.
\newblock Human motion modeling using dvgans.
\newblock {\em arXiv preprint arXiv:1804.10652}, 2018.

\bibitem{liu2022prompt}
Lingbo Liu, Bruce~XB Yu, Jianlong Chang, Qi Tian, and Chang-Wen Chen.
\newblock Prompt-matched semantic segmentation.
\newblock {\em arXiv preprint arXiv:2208.10159}, 2022.

\bibitem{MatthewLoper2015SMPLAS}
Matthew Loper, Naureen Mahmood, Javier Romero, Gerard Pons-Moll, and Michael~J.
  Black.
\newblock Smpl: a skinned multi-person linear model.
\newblock {\em international conference on computer graphics and interactive
  techniques}, 2015.

\bibitem{NaureenMahmood2019AMASSAO}
Naureen Mahmood, Nima Ghorbani, Nikolaus~F. Troje, Gerard Pons-Moll, and
  Michael~J. Black.
\newblock Amass: Archive of motion capture as surface shapes.
\newblock {\em international conference on computer vision}, 2019.

\bibitem{DushyantMehta2016Monocular3H}
Dushyant Mehta, Helge Rhodin, Dan Casas, Pascal Fua, Oleksandr Sotnychenko,
  Weipeng Xu, and Christian Theobalt.
\newblock Monocular 3d human pose estimation in the wild using improved cnn
  supervision.
\newblock {\em international conference on 3d vision}, 2016.

\bibitem{OscarMichel2022Text2MeshTN}
Oscar Michel, Roi Bar-On, Richard Liu, Sagie Benaim, and Rana Hanocka.
\newblock Text2mesh: Text-driven neural stylization for meshes.
\newblock 2022.

\bibitem{olah2017feature}
Chris Olah, Alexander Mordvintsev, and Ludwig Schubert.
\newblock Feature visualization.
\newblock {\em Distill}, 2(11):e7, 2017.

\bibitem{DongHukPark2021BenchmarkFC}
Dong~Huk Park, Samaneh Azadi, Xihui Liu, Trevor Darrell, and Anna Rohrbach.
\newblock Benchmark for compositional text-to-image synthesis.
\newblock {\em neural information processing systems}, 2021.

\bibitem{AdamPaszke2019PyTorchAI}
Adam Paszke, Sam Gross, Francisco Massa, Adam Lerer, James Bradbury, Gregory
  Chanan, Trevor Killeen, Zeming Lin, Natalia Gimelshein, Luca Antiga, Alban
  Desmaison, Andreas Kopf, Edward~Z. Yang, Zachary DeVito, Martin Raison,
  Alykhan Tejani, Sasank Chilamkurthy, Benoit Steiner, Lu Fang, Junjie Bai, and
  Soumith Chintala.
\newblock Pytorch: An imperative style, high-performance deep learning library.
\newblock {\em neural information processing systems}, 2019.

\bibitem{OrPatashnik2021StyleCLIPTM}
Or Patashnik, Zongze Wu, Eli Shechtman, Daniel Cohen-Or, and Dani Lischinski.
\newblock Styleclip: Text-driven manipulation of stylegan imagery.
\newblock {\em arXiv: Computer Vision and Pattern Recognition}, 2021.

\bibitem{SMPL-X:2019}
Georgios Pavlakos, Vasileios Choutas, Nima Ghorbani, Timo Bolkart, Ahmed A.~A.
  Osman, Dimitrios Tzionas, and Michael~J. Black.
\newblock Expressive body capture: 3d hands, face, and body from a single
  image.
\newblock In {\em Proceedings IEEE Conf. on Computer Vision and Pattern
  Recognition (CVPR)}, 2019.

\bibitem{peng2021animatable}
Sida Peng, Junting Dong, Qianqian Wang, Shangzhan Zhang, Qing Shuai, Hujun Bao,
  and Xiaowei Zhou.
\newblock Animatable neural radiance fields for human body modeling.
\newblock 2021.

\bibitem{peng2021neural}
Sida Peng, Yuanqing Zhang, Yinghao Xu, Qianqian Wang, Qing Shuai, Hujun Bao,
  and Xiaowei Zhou.
\newblock Neural body: Implicit neural representations with structured latent
  codes for novel view synthesis of dynamic humans.
\newblock In {\em Proceedings of the IEEE/CVF Conference on Computer Vision and
  Pattern Recognition}, pages 9054--9063, 2021.

\bibitem{MathisPetrovich2021ActionConditioned3H}
Mathis Petrovich, Michael~J. Black, and G{\"u}l Varol.
\newblock Action-conditioned 3d human motion synthesis with transformer vae.
\newblock {\em international conference on computer vision}, 2021.

\bibitem{LeonidPishchulin2017BuildingSS}
Leonid Pishchulin, Stefanie Wuhrer, Thomas Helten, Christian Theobalt, and
  Bernt Schiele.
\newblock Building statistical shape spaces for 3d human modeling.
\newblock {\em Pattern Recognition}, 2017.

\bibitem{MatthiasPlappert2016TheKM}
Matthias Plappert, Christian Mandery, and Tamim Asfour.
\newblock The kit motion-language dataset.
\newblock {\em Big Data}, 2016.

\bibitem{poole2022dreamfusion}
Ben Poole, Ajay Jain, Jonathan~T Barron, and Ben Mildenhall.
\newblock Dreamfusion: Text-to-3d using 2d diffusion.
\newblock {\em arXiv}, 2022.

\bibitem{AbhinandaRPunnakkal2021BABELBA}
Abhinanda~R. Punnakkal, Arjun Chandrasekaran, Nikos Athanasiou, Alejandra
  Quiros-Ramirez, and Michael~J. Black.
\newblock Babel: Bodies, action and behavior with english labels.
\newblock {\em computer vision and pattern recognition}, 2021.

\bibitem{radford2021learning}
Alec Radford, Jong~Wook Kim, Chris Hallacy, Aditya Ramesh, Gabriel Goh,
  Sandhini Agarwal, Girish Sastry, Amanda Askell, Pamela Mishkin, Jack Clark,
  et~al.
\newblock Learning transferable visual models from natural language
  supervision.
\newblock In {\em International Conference on Machine Learning}, pages
  8748--8763. PMLR, 2021.

\bibitem{AlecRadford2021LearningTV}
Alec Radford, Jong~Wook Kim, Chris Hallacy, Aditya Ramesh, Gabriel Goh,
  Sandhini Agarwal, Girish Sastry, Amanda Askell, Pamela Mishkin, Jack Clark,
  Gretchen Krueger, and Ilya Sutskever.
\newblock Learning transferable visual models from natural language
  supervision.
\newblock {\em international conference on machine learning}, 2021.

\bibitem{RobinRombach2022HighResolutionIS}
Robin Rombach, Andreas Blattmann, Dominik Lorenz, Patrick Esser, and Bj\\"orn
  Ommer.
\newblock High-resolution image synthesis with latent diffusion models.
\newblock 2022.

\bibitem{ChitwanSaharia2022PhotorealisticTD}
Chitwan Saharia, William Chan, Saurabh Saxena, Lala Li, Jay Whang, Emily
  Denton, Seyed Kamyar, Seyed Ghasemipour, Burcu Karagol, S~Sara Mahdavi,
  Rapha~Gontijo Lopes, Tim Salimans, Jonathan Ho, David~J Fleet, and Mohammad
  Norouzi.
\newblock Photorealistic text-to-image diffusion models with deep language
  understanding.
\newblock 2022.

\bibitem{AdityaSanghi2021CLIPForgeTZ}
Aditya Sanghi, Hang Chu, Joseph~G. Lambourne, Ye Wang, Chin-Yi Cheng, and Marco
  Fumero.
\newblock Clip-forge: Towards zero-shot text-to-shape generation.
\newblock {\em arXiv: Computer Vision and Pattern Recognition}, 2021.

\bibitem{singer2022make}
Uriel Singer, Adam Polyak, Thomas Hayes, Xi Yin, Jie An, Songyang Zhang, Qiyuan
  Hu, Harry Yang, Oron Ashual, Oran Gafni, et~al.
\newblock Make-a-video: Text-to-video generation without text-video data.
\newblock {\em arXiv preprint arXiv:2209.14792}, 2022.

\bibitem{sun2022paradigm}
Tian-Xiang Sun, Xiang-Yang Liu, Xi-Peng Qiu, and Xuan-Jing Huang.
\newblock Paradigm shift in natural language processing.
\newblock {\em Machine Intelligence Research}, 19(3):169--183, 2022.

\bibitem{tevet2022motionclip}
Guy Tevet, Brian Gordon, Amir Hertz, Amit~H Bermano, and Daniel Cohen-Or.
\newblock Motionclip: Exposing human motion generation to clip space.
\newblock {\em arXiv preprint arXiv:2203.08063}, 2022.

\bibitem{GlVarol2017LearningFS}
G{\"u}l Varol, Javier Romero, Xavier Martin, Naureen Mahmood, Michael~J. Black,
  Ivan Laptev, and Cordelia Schmid.
\newblock Learning from synthetic humans.
\newblock {\em computer vision and pattern recognition}, 2017.

\bibitem{vaswani2017attention}
Ashish Vaswani, Noam Shazeer, Niki Parmar, Jakob Uszkoreit, Llion Jones,
  Aidan~N Gomez, {\L}ukasz Kaiser, and Illia Polosukhin.
\newblock Attention is all you need.
\newblock {\em Advances in neural information processing systems}, 30, 2017.

\bibitem{villegas2022phenaki}
Ruben Villegas, Mohammad Babaeizadeh, Pieter-Jan Kindermans, Hernan Moraldo,
  Han Zhang, Mohammad~Taghi Saffar, Santiago Castro, Julius Kunze, and Dumitru
  Erhan.
\newblock Phenaki: Variable length video generation from open domain textual
  description.
\newblock {\em arXiv preprint arXiv:2210.02399}, 2022.

\bibitem{YaelVinker2022CLIPassoSO}
Yael Vinker, Ehsan Pajouheshgar, Jessica~Y. Bo, Roman~Christian Bachmann,
  Amit~Haim Bermano, Daniel Cohen-Or, Amir Zamir, and Ariel Shamir.
\newblock Clipasso: Semantically-aware object sketching.
\newblock 2022.

\bibitem{TimovonMarcard2018RecoveringA}
Timo von Marcard, Roberto Henschel, Michael~J. Black, Bodo Rosenhahn, and
  Gerard Pons-Moll.
\newblock Recovering accurate \{3D\} human pose in the wild using \{IMUs\} and
  a moving camera.
\newblock {\em european conference on computer vision}, 2018.

\bibitem{wang2023lion}
Haixin Wang, Jianlong Chang, Xiao Luo, Jinan Sun, Zhouchen Lin, and Qi Tian.
\newblock Lion: Implicit vision prompt tuning, 2023.

\bibitem{wang2022fine}
Shijie Wang, Jianlong Chang, Zhihui Wang, Haojie Li, Wanli Ouyang, and Qi Tian.
\newblock Fine-grained retrieval prompt tuning.
\newblock {\em arXiv preprint arXiv:2207.14465}, 2022.

\bibitem{youwang2022clip}
Kim Youwang, Kim Ji-Yeon, and Tae-Hyun Oh.
\newblock Clip-actor: Text-driven recommendation and stylization for animating
  human meshes.
\newblock {\em arXiv preprint arXiv:2206.04382}, 2022.

\bibitem{yu2022towards}
Bruce~XB Yu, Jianlong Chang, Lingbo Liu, Qi Tian, and Chang~Wen Chen.
\newblock Towards a unified view on visual parameter-efficient transfer
  learning.
\newblock {\em arXiv preprint arXiv:2210.00788}, 2022.

\bibitem{zhang2022motiondiffuse}
Mingyuan Zhang, Zhongang Cai, Liang Pan, Fangzhou Hong, Xinying Guo, Lei Yang,
  and Ziwei Liu.
\newblock Motiondiffuse: Text-driven human motion generation with diffusion
  model.
\newblock {\em arXiv preprint arXiv:2208.15001}, 2022.

\bibitem{zhou2019continuity}
Yi Zhou, Connelly Barnes, Jingwan Lu, Jimei Yang, and Hao Li.
\newblock On the continuity of rotation representations in neural networks.
\newblock In {\em Proceedings of the IEEE/CVF Conference on Computer Vision and
  Pattern Recognition}, pages 5745--5753, 2019.

\end{thebibliography}
}

\end{document}